\documentclass[letterpaper, 10 pt, conference]{ieeeconf}  

\IEEEoverridecommandlockouts                              

\usepackage{bm}
\usepackage{url}
\usepackage{graphicx}
\usepackage{lipsum}
\usepackage{amsmath}
\usepackage{amssymb}
\usepackage{nccmath}
\usepackage{comment}
\usepackage{balance}
\usepackage{multirow}
\usepackage{textcomp}
\usepackage{multirow}
\usepackage{subcaption}
\usepackage{booktabs}
\usepackage{algorithm}
\usepackage{dblfloatfix}
\usepackage[table]{xcolor}
\usepackage[noend]{algpseudocode}

\title{\LARGE \bf
Scalable Fiducial Tag Localization on a 3D Prior Map \\ via Graph-Theoretic Global Tag-Map Registration
}

\author{Kenji Koide$^{1}$, Shuji Oishi$^{1}$, Masashi Yokozuka$^{1}$, and Atsuhiko Banno$^{1}$
\thanks{*This work was supported in part by a project commissioned by the New Energy and Industrial Technology Development Organization (NEDO).}
\thanks{$^{1}$All the authors are with the Department of Information Technology and Human Factors, the National Institute of Advanced Industrial Science and Technology, Umezono 1-1-1, Tsukuba, 3050061, Ibaraki, Japan, {\tt\small k.koide@aist.go.jp}}%
}

\begin{document}

\maketitle
\thispagestyle{empty}
\pagestyle{empty}

\begin{abstract}

This paper presents an accurate and scalable method for fiducial tag localization on a 3D prior environmental map. The proposed method comprises three steps: 1) visual odometry-based landmark SLAM for estimating the relative poses between fiducial tags, 2) geometrical matching-based global tag-map registration via maximum clique finding, and 3) tag pose refinement based on direct camera-map alignment with normalized information distance. Through simulation-based evaluations, the proposed method achieved a 98 \% global tag-map registration success rate and an average tag pose estimation accuracy of a few centimeters. Experimental results in a real environment demonstrated that it enables to localize over 110 fiducial tags placed in an environment in 25 minutes for data recording and post-processing.

\end{abstract}

\section{Introduction}

In recent years, map-based visual localization methods have been actively studied and widely used for autonomous navigation systems \cite{Caselitz2016,Oishi2020} and user interaction applications (e.g., augmented reality \cite{Park2012}). These methods employing precise 3D maps enable accurate localization and navigation in a large variety of environments with affordable equipment. These visual localization methods are, however, still error-prone in feature-less and dynamic environments, and it is sometimes necessary to rely on visual fiducial tags \cite{Krogius2019,Huang2021} for initialization and fail-safe. Deploying a number of fiducial tags in the environment and combining them with visual localization methods allows us to build a robust and accurate localization and navigation system \cite{Huang2018,fangmarker,Kayhani2022}.

Deploying many fiducial tags on a 3D prior map is, however, sometimes difficult and tedious. Because the tag localization accuracy directly affects the navigation accuracy, we need to precisely determine the tag positions as accurate as possible. Furthermore, we need to place many fiducial tags (several hundreds, possibly) to cover the entire environment. However, fiducial tag positions on a prior map are often measured by hand in many works, which results in large human effort and inaccurate localization.

\begin{figure}[tb]
    \centering
    \includegraphics[width=\linewidth]{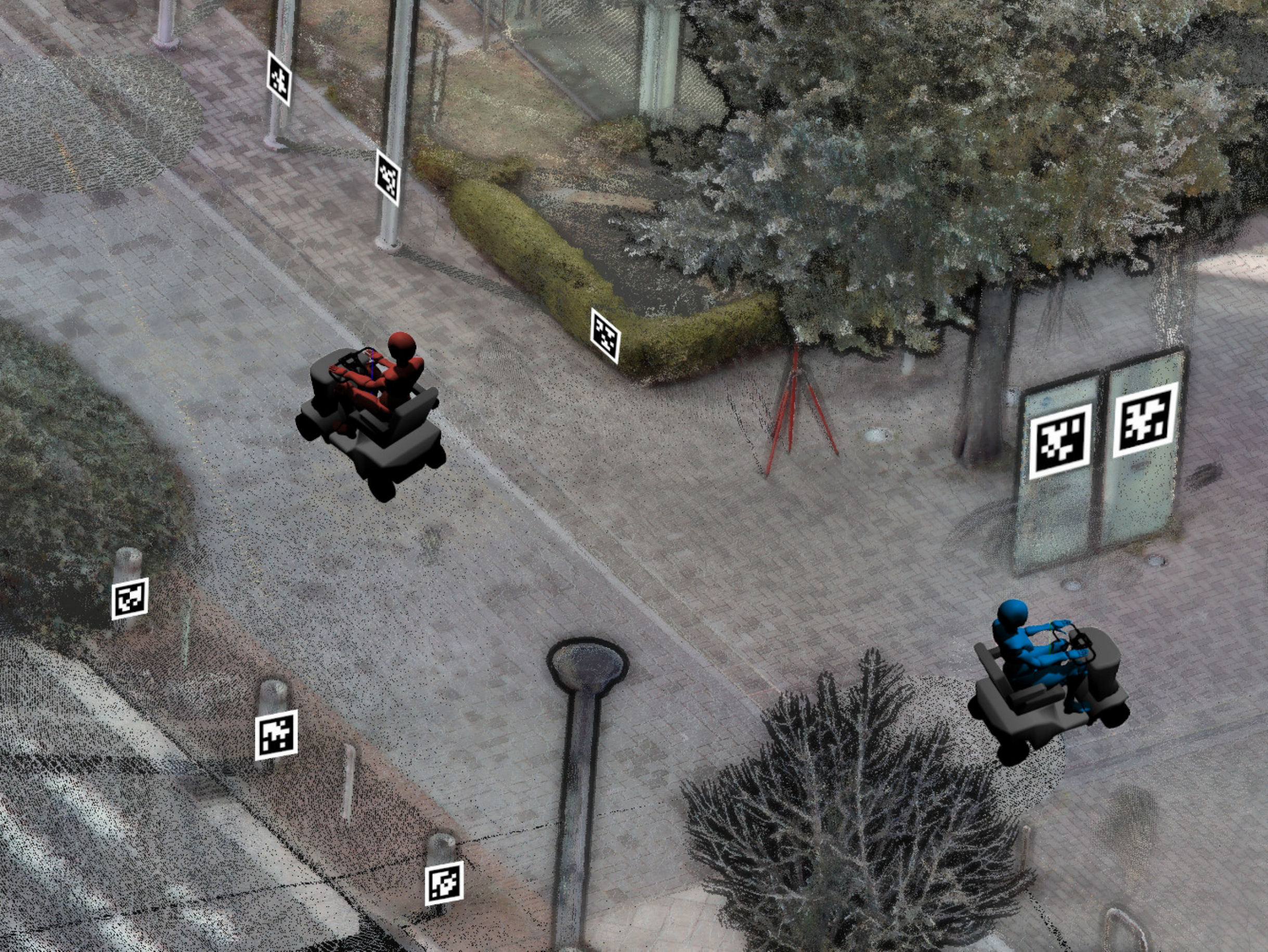}
    \caption{With the proposed fiducial tag localization method, we aim to make it easy to develop a system based on robust vision-based localization using precisely localized tags on a precise 3D environmental map.}
    \label{fig:tags_on_prior_map}
\end{figure}

\begin{figure*}[tb]
    \centering
    \includegraphics[width=\linewidth]{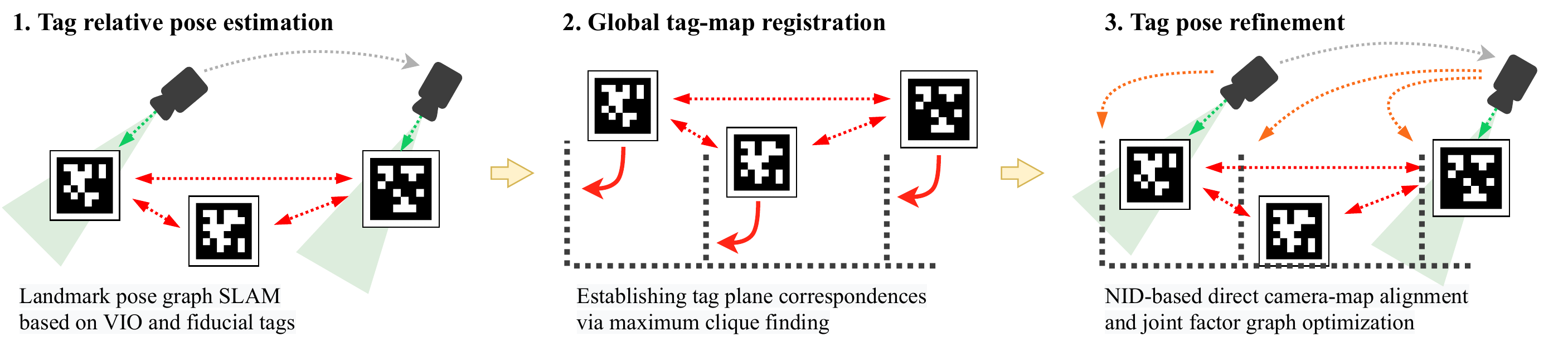}
    \caption{Processing flow of the proposed system.}
    \label{fig:system}
\end{figure*}

In this work, we propose an automatic method for fiducial tag localization on a 3D prior map. The proposed method enables to accurately determine the poses of many fiducial tags on a 3D prior map in a short time (e.g., more than 100 tags in less than 25 minutes) as a pre-installation process of automatic navigation and user interaction systems. We consider utilizing the precisely localized fiducial tags makes it easy to build a system based on vision-based localization like that shown in Fig. \ref{fig:tags_on_prior_map}.

The proposed method comprises three steps: 1) We first estimate the relative poses between fiducial tags via visual inertial odometry (VIO)-based landmark pose graph SLAM. We observe each fiducial tag with an agile camera and construct a pose graph in which fiducial tag poses are bridged by VIO trajectory edges. 2) We then roughly align the fiducial tags with a 3D prior map (i.e., global tag-map registration). Inspired by the recent graph-theoretic global registration methods \cite{Yang2021,Shi2021,Lusk2021}, we propose a tag-map matching method based on robust tag-plane correspondence estimation via maximum clique finding. 3) Finally, we refine the estimated tag poses by directly aligning agile camera images with the prior map using normalized information distance (NID), a mutual-information-based cross-modal distance metric.

The main contribution of this work is three-fold:
\begin{enumerate}
    \item We propose an accurate and scalable fiducial tag localization method that enables deploying a massive amount of tags on a 3D prior map in a short time.
    \item To robustly perform global tag-map registration, a graph-theoretic tag-plane correspondence estimation method is proposed.
    \item We show that the combination of NID-based direct camera-map alignment and maximum clique finding-based outlier filtering enables to further improve the fiducial tag localization accuracy.
\end{enumerate}

\section{Related work}

There have been proposed many monocular camera localization methods in a 3D prior map for vision-based navigation. Caselitz et al. reconstructed the surrounding environment from camera images using a visual SLAM technique and estimated the camera pose in a given map by matching reconstructed points with map points \cite{Caselitz2016}. Pascoe et al., used the NID metric, a mutual-information-based cross-modal distance metric, to directly align camera images with the 3D prior map \cite{Pascoe2015}. Ye et al., combined surfel-based map rendering and direct photometric comparison to keep tracking the camera trajectory on a prior map \cite{Ye2020}. While these methods enable accurate map-referenced camera localization and affordable vision-based navigation with a single camera, they can sometimes be unreliable in feature-less and dynamic environments. Several practical systems thus often combine vision-based camera localization with fiducial tag detection for reliability and for fail-safe \cite{Huang2018, Eckenhoff2019, beul2018fast}.

As we can robustly detect fiducial tags on an image, by using a visual odometry technique, their poses with respect to the visual odometry reference frame can easily be estimated in the form of the landmark pose graph SLAM \cite{pfrommer2019tagslam,Koide2020}. However, aligning the estimated tag poses with a 3D prior map (i.e., tag-map global registration) is not straightforward because of the difference of modalities between visually detected fiducial tags and a 3D point cloud map. The modality difference makes it difficult to apply image-to-image matching methods \cite{GalvezLopez2012,Cattaneo2020} nor geometry-based global registration methods \cite{Yang2021,Buch2013} to estimate the transformation between tag and map reference frames. 

If a 3D prior map is recorded in an ordered point cloud format (e.g., PTX format), it would be possible to generate images from points and perform visual image matching (e.g., \cite{Schonberger2016}). However, many 3D map datasets provide only unordered point clouds that make it difficult to generate images with good quality resulting in deteriorated accuracy of visual image matching.

The proposed method robustly determines the tag-map transformation across different modalities by combining geometry-based tag-plane correspondence hypothesis making and graph-theoretic outlier hypothesis rejection. While the geometrical hypothesis making yields many false correspondences, the graph-theoretic algorithm robustly filters out wrong hypotheses and finds the best subset of correspondence hypotheses that gives the best explanation for the tag placement in the map.

\section{Methodology}

Fig. \ref{fig:system} shows an overview of the proposed method. In the tag relative pose estimation step, we observe each fiducial tag using an agile camera and estimate the relative poses between fiducial tags in the form of the landmark visual SLAM. In the following step, we roughly align fiducial tags with a 3D prior map by establishing tag-plane correspondences via maximum clique finding. We then refine tag and camera poses by directly aligning camera images with the map.

\subsection{Tag Relative Pose Estimation based on Landmark Pose Graph SLAM}

In this step, we use an agile camera to observe fiducial tags placed in an environment, and reconstruct the relative poses between tags with a standard landmark pose graph SLAM approach \cite{Grisetti2010}. We estimate the camera ego-motion using a VIO algorithm (e.g., VINS-Mono \cite{Qin2018}) while detecting fiducial tags on images \cite{Wang2016}. Let $T_{WC} (t)$ be the camera pose estimated by VIO at time $t$, and $T_{CT}^i (t)$ be the pose of a detected fiducial tag with tag ID $ = i$. We estimate the camera pose at every time step $\widetilde{T}_{WC}(t)$ and fiducial tag poses $\widetilde{T}_{WT}^i$ by minimizing the following objective function that combines odometry factors $e_{\text odom}$ and tag observation factors $e_{\text tag}$:

\begin{align}
e_{\text{slam}} &= e_{\text{odom}} + e_{\text{tag}}, \\
e_{\text{odom}} &= \sum_{t=1}^{N - 1} \| \log \left( \widetilde{T}_{WC}(t, t+1)^{-1} T_{WC}(t, t+1)  \right) \|^2, \\
e_{\text{tag}} &= \sum_{t=1}^{N} \sum_{i}^M \| \log \left( \widetilde{T}_{WC}(t)^{-1} \widetilde{T}_{WM}^i T_{CM}^i(t)^{-1} \right) \|^2,
\end{align}
where $T_{WC}(t, t+1) = T_{WC}(t)^{-1} T_{WC}(t)$ is the relative camera pose between $t$ and $t+1$, and $\log$ is the SE3 logarithmic map. Here, we intentionally decouple VIO estimation and tag pose estimation and fuse them on a pose graph so that we can easily change the VIO algorithm depending on the use scenario (e.g., using another VIO running on a smartphone \cite{MarderEppstein2016} instead of VINS-Mono).

\subsection{Tag-Map Global Registration via Graph-Theoretic Tag-Plane Correspondence Establishment}

Given the estimated relative poses between fiducial tags, we roughly align the reference frame of the tags (i.e., VIO origin) with the map reference frame (i.e., tag-map global registration). The challenge here is that we need to deal with the difference of modalities between the sparse point cloud map and visually detected fiducial tags. Due to the difference of modalities, traditional vision-based image matching methods \cite{GalvezLopez2012,Cattaneo2020} are not applicable.

In this work, we assume that most of the fiducial tags are placed on a plane in the environment, and solve the global registration problem by establishing tag-plane correspondences. This assumption can naturally be held in many practical use scenarios because most fiducial tags are required to be placed on a flat surface for accurate detection and localization \cite{Wang2016}.

\begin{figure}[tb]
    \centering
    \includegraphics[width=\linewidth]{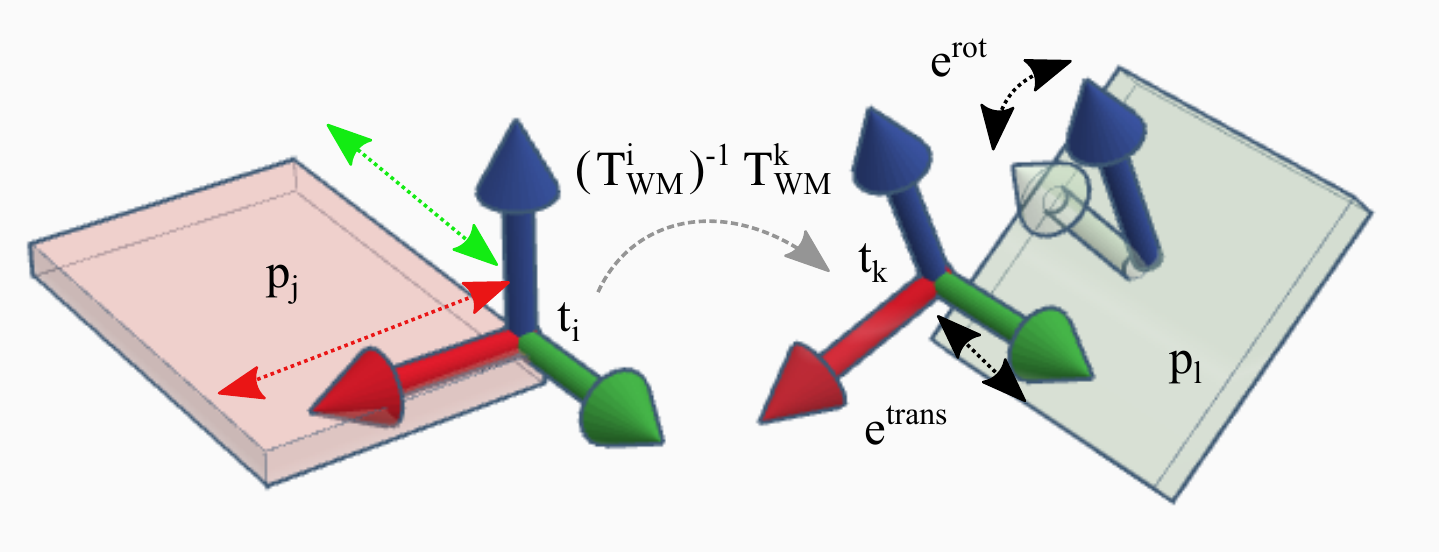}
    \caption{Tag-plane correspondence consistency.}
    \label{fig:tag_plane}
\end{figure}

Inspired by the recent success of graph-theoretic approaches for global registration \cite{Yang2021,Shi2021,Lusk2021}, we robustly estimate tag-plane correspondences via maximum clique finding.  We first construct a consistency graph, in which vertices represent a hypothesis of tag-plane correspondence, and edges represent the consistency between two tag-plane correspondence hypotheses. By finding the largest subset of hypotheses that are all mutually consistent (i.e., the maximum clique in the consistency graph), we can robustly filter out outlier correspondences and determine a set of tag-plane correspondences that gives the best explanation for the tag placement in the prior map.

To construct a consistency graph, we first extract planes from the map point cloud using region growing segmentation \cite{rabbani2006segmentation} and then list all possible tag-plane correspondences. We evaluate the geometrical consistency of every combination of correspondence hypotheses as illustrated in Fig. \ref{fig:tag_plane}. Let $h_{ij} = (t_i, p_j)$ be a hypothesis that a fiducial tag $t_i$ corresponds to a plane $p_j$ in the map and consider its consistency with another hypothesis $h_{kl} = (t_k, p_l)$. We transform the pose of $t_i$ such that its normal becomes aligned with the normal of $p_j$. Given the relative pose between $t_i$ and $t_k$, we shift and rotate $t_i$ on the plane $p_j$ such that the distance between $t_k$ and $p_l$ is minimized. If the distance between $t_k$ and $p_l$ is smaller than $\text{th}^\text{trans}$ and the angle error of their normals is smaller than $\text{th}^\text{rot}$ (e.g., 0.4 m and 10\textdegree), we consider $h_{ij}$ and $h_{kl}$ are mutually consistent. For the algorithmic detail of the consistency check procedure, see the appendix. 

With all combinations of tag-plane correspondence hypotheses that pass the mutual geometrical consistency check, we construct a consistency graph and find the maximum clique using the parallel heuristic maximum clique finding algorithm \cite{Rossi2015}.

\begin{figure}[tb]
  \centering
  \begin{minipage}[b]{\linewidth}
  \centering
  \includegraphics[width=\linewidth]{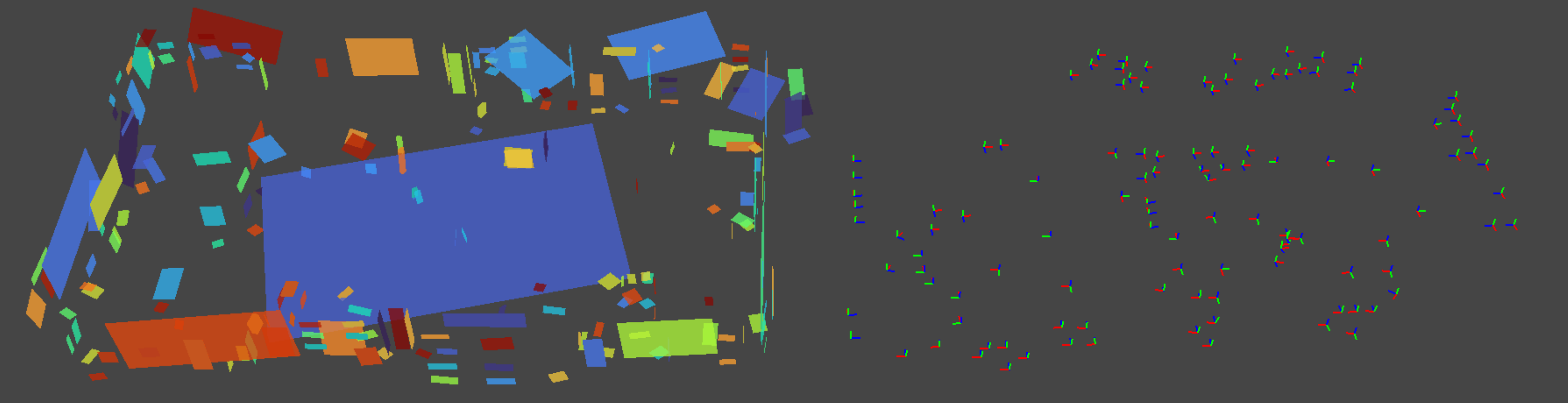}
  \subcaption{Planes in a map (left) and fiducial tag poses (right)}
  \end{minipage}
  \begin{minipage}[b]{\linewidth}
  \centering
  \includegraphics[width=\linewidth]{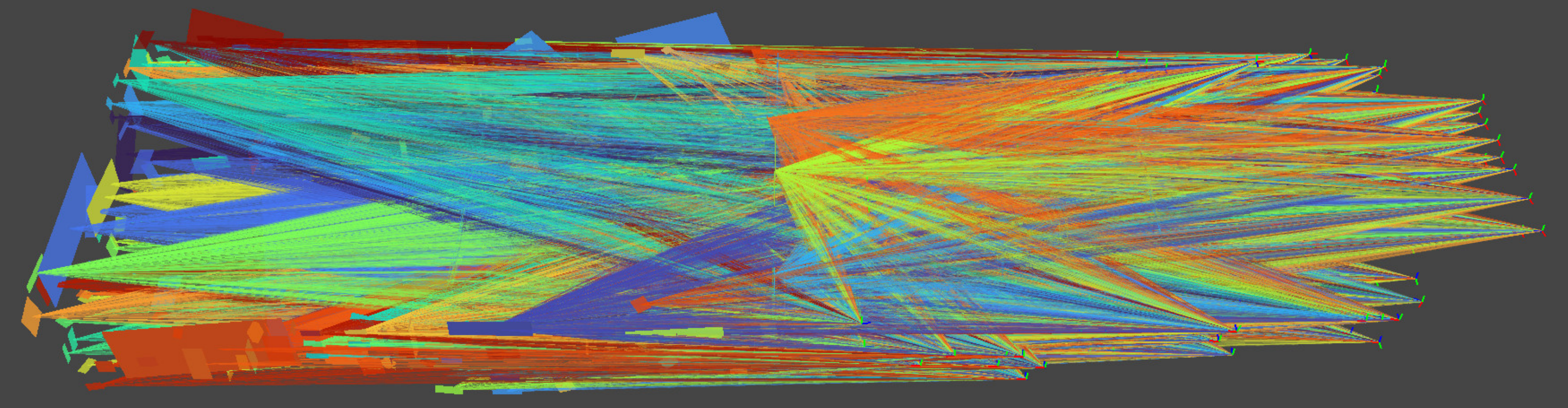}
  \subcaption{Consistency graph (tag-plane correspondence hypotheses)}
  \end{minipage}
  \begin{minipage}[b]{\linewidth}
  \centering
  \includegraphics[width=\linewidth]{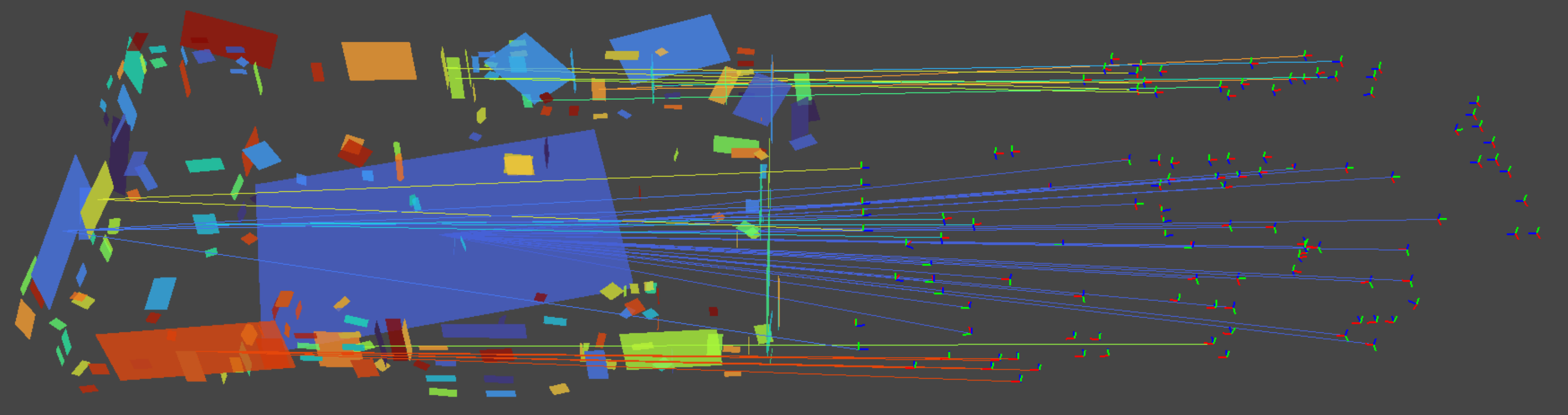}
  \subcaption{Maximum clique in the consistency graph}
  \end{minipage}
  \caption{Tag-plane correspondence estimation via maximum clique finding.}
  \label{fig:max_clique}
\end{figure}

Fig. \ref{fig:max_clique} shows an example of tag-plane correspondence estimation results. For all possible combinations of planes and tags, we evaluated the correspondence consistency and constructed a consistency graph (Fig. \ref{fig:max_clique} (b)) and then found the maximum clique (Fig. \ref{fig:max_clique} (c)). While the consistency graph contained a massive amount of tag-plane correspondence hypotheses (429,735 hypothesis pairs), the maximum clique (56 tag-plane correspondences) was efficiently found in 92 msec. 

Given the tag-plane correspondences, we estimate the transformation between the tag and map reference frames by minimizing the symmetric point-to-normal ICP distance \cite{Rusinkiewicz2019} between corresponding tags and planes.

\subsection{Estimation Refinement via Information-theoretic Direct Camera-Map Alignment}

We then refine the tag and camera pose estimates by directly aligning each camera image with the global map using the NID \cite{Pascoe2015}, a mutual-information-based cross-modal distance measure. We insert the camera poses aligned with the map into the landmark pose graph as prior factors, and re-optimize the graph to improve the camera and tag pose estimates.

Let $\mathcal{P}$ be a map point cloud, $I_r$ be a camera image, and $\widetilde{T}_{WC}(t)$ be the camera pose with respect to the map reference frame. Given an initial estimate of $\widetilde{T}_{WC}(t)$, we first remove points from $\mathcal{P}$ that should not be visible from the current viewpoint using direct visibility assessment \cite{Katz2007}, and then estimate $\widetilde{T}_{WC}(t)$ by minimizing the NID metric between $\mathcal{P}$ and $I_r$ using the BFGS algorithm. The NID is defined as follows:

\begin{align}
\label{eq:nid}
\text{NID}(I_r, I_s) &= \frac{\text{H}(I_r, I_s) - \text{MI}(I_r; I_s)}{\text{H}(I_r, I_s)}, \\
\text{MI}(I_r; I_s) &= \text{H}(I_r) + \text{H}(I_s) - \text{H}(I_r, I_s),
\end{align}
where $I_s$ is a map image created by projecting $\mathcal{P}$ on the image space of $I_r$, $\text{H}(I_r, I_s), \text{H}(I_r), \text{H}(I_s)$ are the joint and marginal entropies of $I_r$ and $I_s$, and $\text{MI}(I_r, I_s)$ is the mutual information between $I_r$ and $I_s$. Because this metric does not directly compare pixel and point colors but measure co-occurrence of them, it enables to measure the distance between data across different modalities. Following \cite{Pascoe2015}, we use B-spline based weighted histogram voting to make Eq. \ref{eq:nid} differentiable.

The NID enables to accurately determine the camera pose with respect to a map point cloud. It is, however, very sensitive to the initial guess and often gets corrupted. Because the NID is a dimensionless quantity, it is not easy to remove corrupted results with simple thresholding of the NID value. 

To robustly filter out corrupted camera-map alignment results, we again use a graph-theoretic approach to find the maximum mutually consistent subset of them. Let $h_t = ( \widetilde{T}_{WC}(t), \widehat{T}_{WC}(t) )$ be a pair of initial and refined camera poses. To determine the consistency between $h_t$ and $h_k$, we calculate the camera pose displacements they are declaring:

\begin{align}
\Delta \widehat{T}_{WC}(t) &= \widetilde{T}_{WC}^{-1}(t) \widehat{T}_{WC}(t), \\
\Delta \widehat{T}_{WC}(t, k) &= \Delta \widehat{T}_{WC}(t)^{-1} \Delta \widehat{T}_{WC}(k).
\end{align}
If the translational and rotational errors of $\Delta \widehat{T}_{WC}(t, k)$ are smaller than threshold values (e.g., 0.5 m and 5\textdegree), we consider $h_t$ and $h_k$ are mutually consistent. We construct a consistency graph for all combinations of camera-map alignment results and find the maximum clique to filter out corrupted results (i.e., outliers).

\begin{figure}[tb]
  \centering
  \begin{minipage}[b]{0.48\linewidth}
  \centering
  \includegraphics[height=5.0cm]{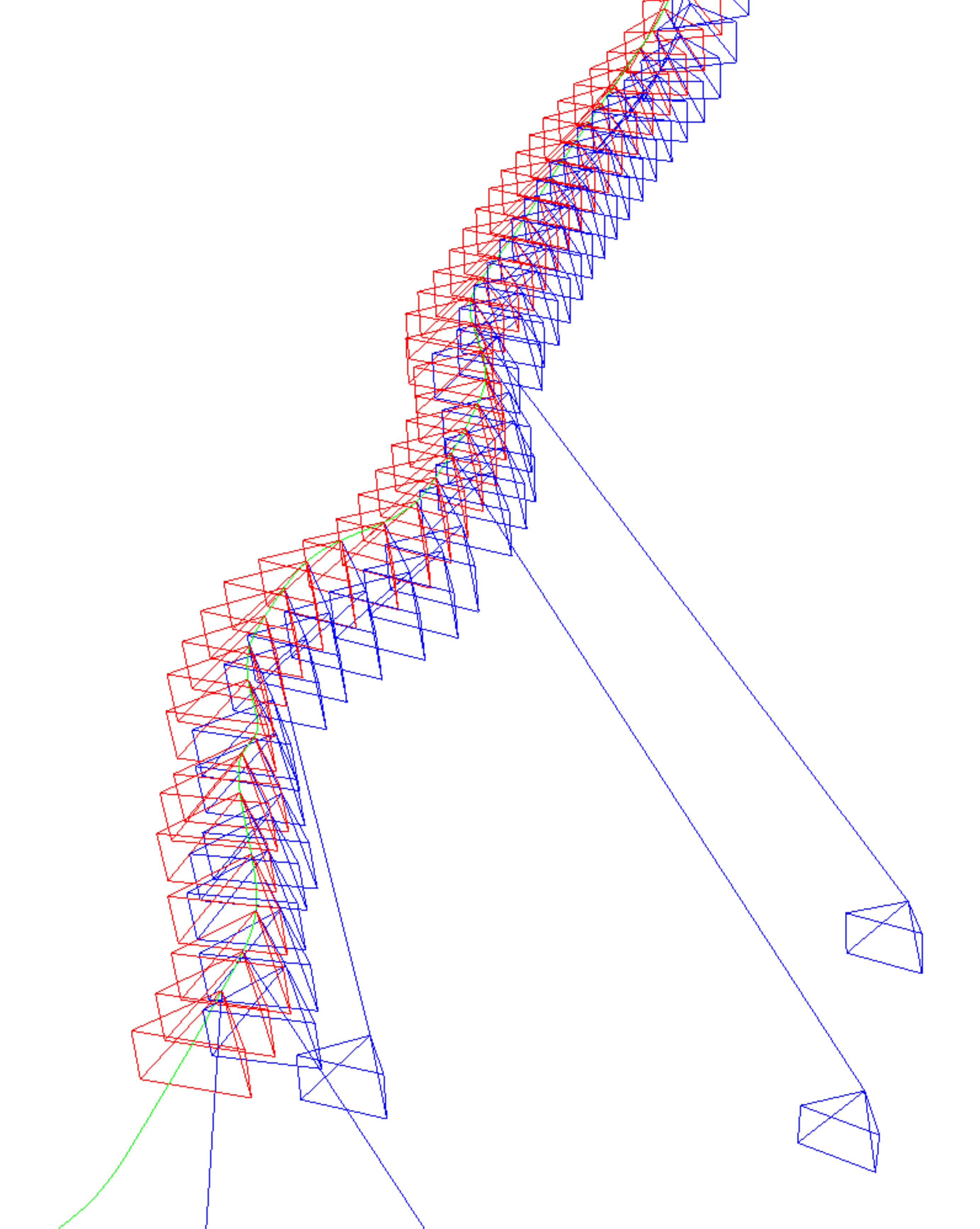}
  \subcaption{Camera alignment results}
  \end{minipage}
  \begin{minipage}[b]{0.48\linewidth}
  \centering
  \includegraphics[height=5.0cm]{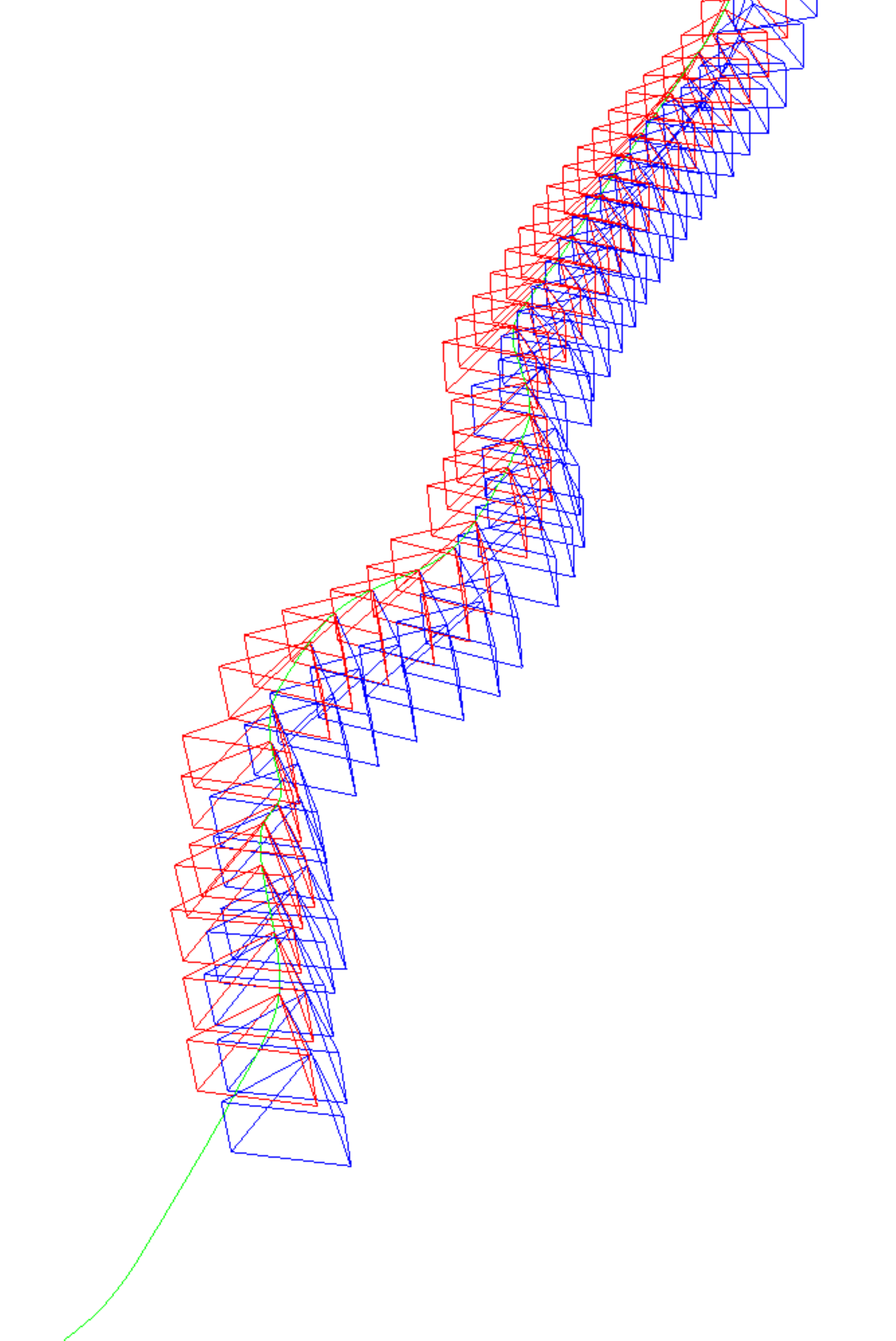}
  \subcaption{Outlier filtering result}
  \end{minipage}
  \caption{Camera-map alignment and outlier filtering results. Red: initial camera poses, Blue: refined camera poses.}
  \label{fig:prior_filter}
\end{figure}

Fig. \ref{fig:prior_filter} (a) shows an example of NID-based camera-map alignment results. Red frustums show the initial camera poses of frames where the BFGS optimization converges while blue ones show estimated camera poses. We can see that the NID-based optimization sometimes gets corrupted. Fig. \ref{fig:prior_filter} (b) shows a result of outlier filtering, in which corrupted camera-map alignment results are filtered out and only mutually consistent results remain. 

We insert the refined camera poses in the factor graph created in the tag relative pose estimation as pose prior factors, and re-optimize tag and camera poses with all the constraints:

\begin{align}
e_{\text{refine}} &= e_{\text{odom}} + e_{\text{tag}} + e_{\text{NID}}, \\
e_{\text{NID}} &= \sum_i^N \| \log \left( \widetilde{T}_{WC}(t)^{-1} \widehat{T}_{WC}(t) \right) \|^2.
\end{align}

\section{Experiments}

\subsection{Evaluation in a Simulated Environment}

\subsubsection{Global registration evaluation}

\begin{figure}[tb]
    \centering
    \includegraphics[width=0.9\linewidth]{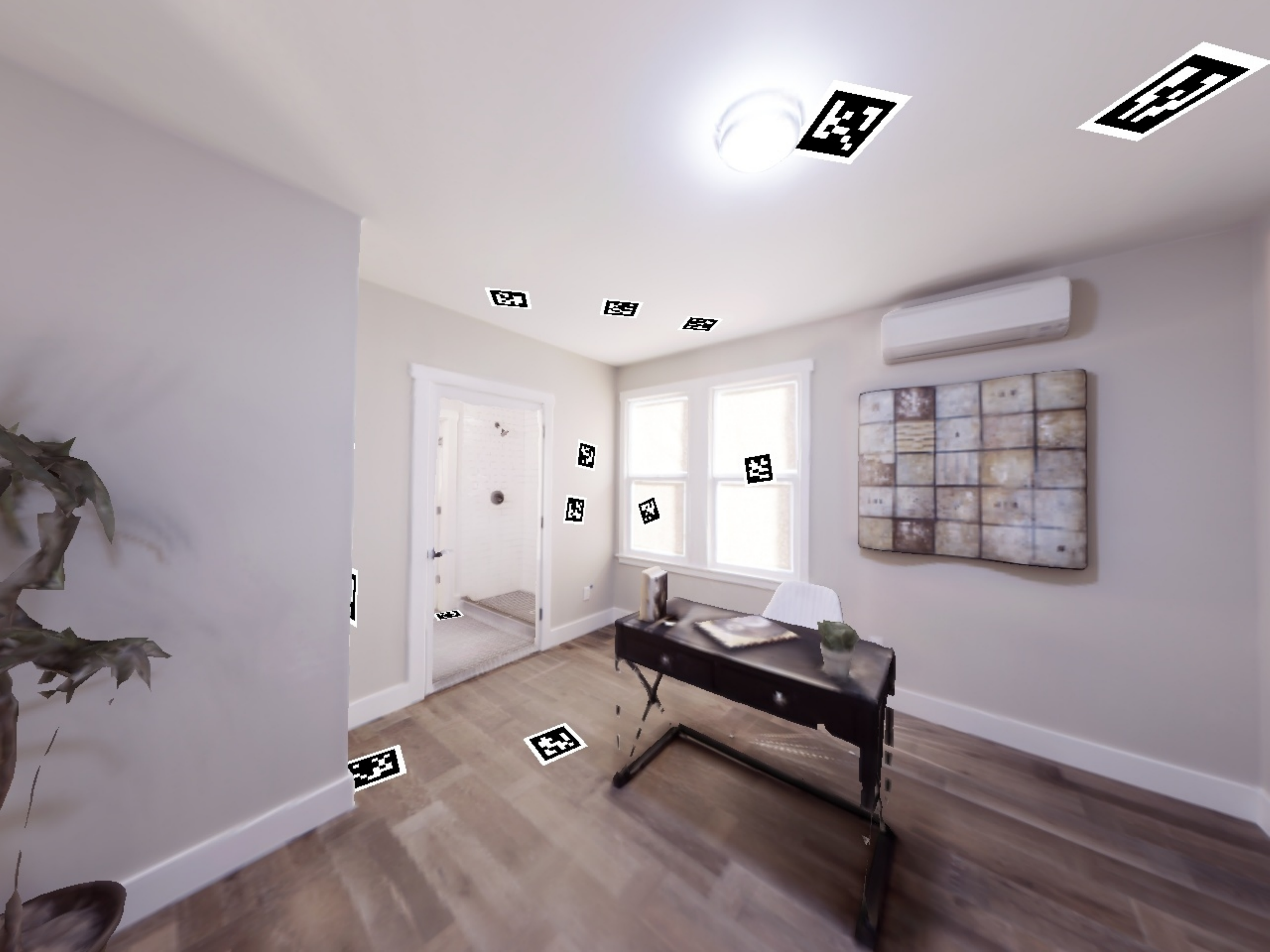}
    \caption{Fiducial tags randomly placed on the Replica dataset.}
    \label{fig:replica_tags}
\end{figure}



\begin{table}[tb]
  \centering
  \caption{Global registration success rate}
  \label{tab:result_global}
  \begin{tabular}{c|ccc}
  \toprule
  Method       & RANSAC \cite{Buch2013} & Teaser \cite{Yang2021}  & Proposed   \\ \midrule
  Success rate & 26\% (13 / 50)         & 70\% (35 / 50)          & 98\% (49 / 50) \\
  \bottomrule
  \end{tabular}
\end{table}

To evaluate the proposed method, we used {\it apartment\_0} model in the {\it Replica} dataset \cite{straub2019replica, Savva2019}. We generated 50 camera trajectories by randomly sampling waypoints in the map and interpolating them using SE3 B-spline interpolation. Along with camera images, we synthesized IMU data using \cite{Geneva2020}. For each trajectory, we randomly placed 200 fiducial tags on planes in the map and estimated the poses of fiducial tags, which were observed by the camera more than once, using the proposed method. 

As a baseline, we compared the proposed method with feature-based global registration methods. We first ran colmap \cite{Schonberger2016} on the camera image stream to obtain a dense 3D point cloud of the environment. We then extracted FPFH features \cite{Rusu2009} respectively from the reconstructed point cloud and the 3D prior model and estimated the transformation between them using RANSAC \cite{Buch2013} and Teaser \cite{Yang2021}. 

\begin{figure}[tb]
  \centering
  \begin{minipage}[b]{0.48\linewidth}
  \centering
  \includegraphics[height=3.5cm]{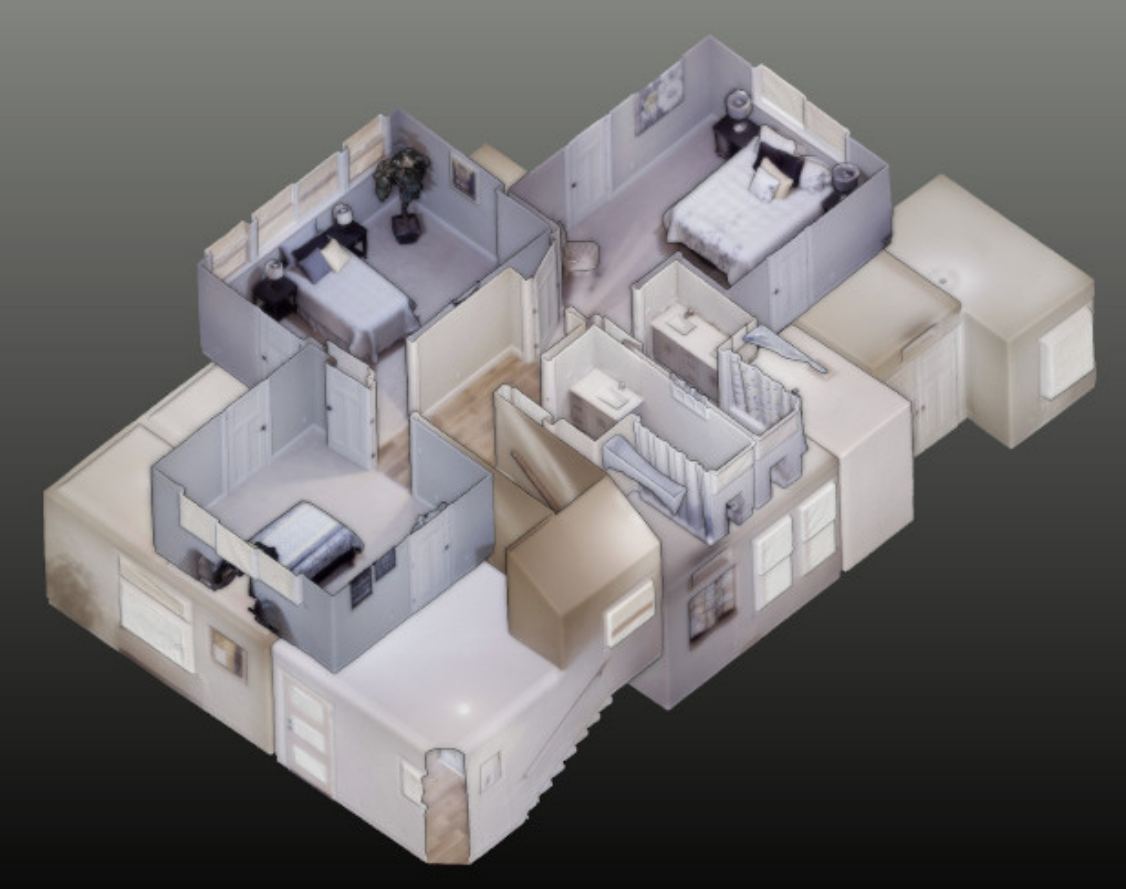}
  \subcaption{Replica model}
  \end{minipage}
  \begin{minipage}[b]{0.48\linewidth}
  \centering
  \includegraphics[height=3.5cm]{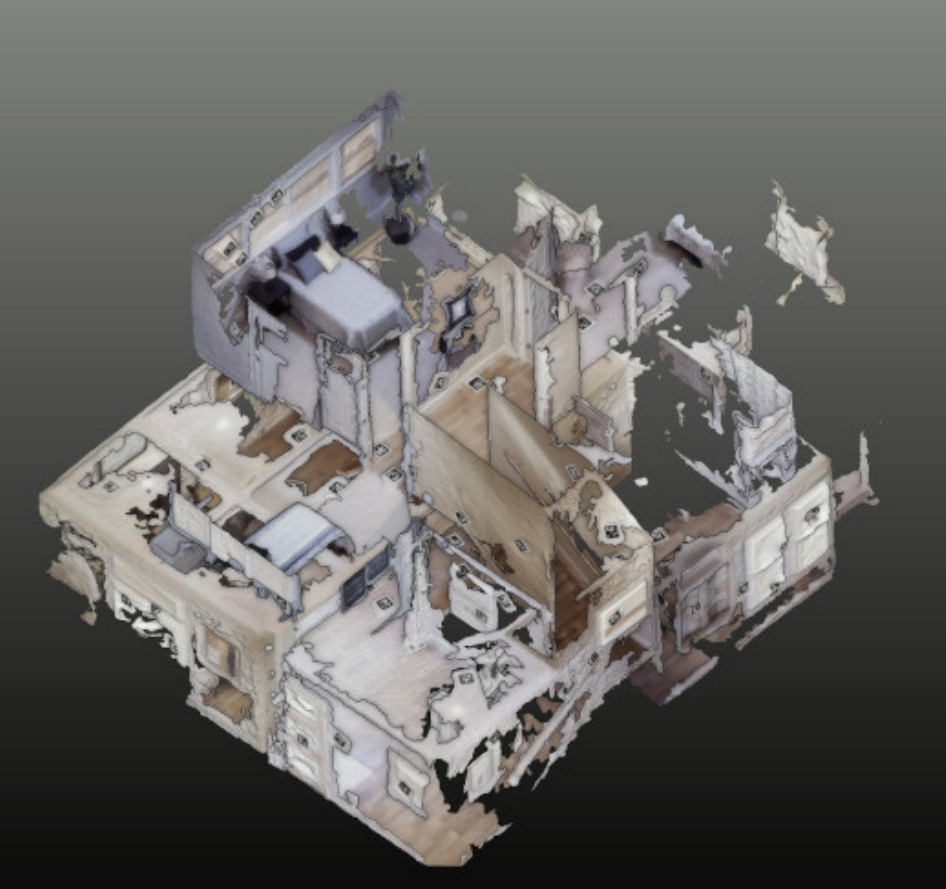}
  \subcaption{Reconstructed model}
  \end{minipage}

  \caption{Prior environmental map and reconstructed point cloud for {\it apartment\_0} model.}
  \label{fig:replica_models}
\end{figure}

We first evaluated the global registration success rate of each method. If the translational and rotational errors of a global registration result are smaller than threshold values (1.0 m and 15\textdegree), we consider the registration is succeeded. Table \ref{tab:result_global} shows the success rate of each global registration method. We can see that, even with Teaser \cite{Yang2021}, a state-of-the-art transformation estimation algorithm, the global registration failed for 30 \% of the sequences. Fig. \ref{fig:replica_models} shows the 3D prior map model and the reconstructed model. While the reconstructed model well captures the overall shape of the environment, the detailed shapes and densities of points are largely different from those of the prior map model. We consider that these differences make it difficult to obtain consistent features between the reconstructed and the prior models, which results in the global registration failures.

The proposed tag-plane matching method successfully estimated the tag-map transformation except for a sequence where fiducial tags were placed very symmetrically and wrong tag-plane correspondences were given via maximum clique finding.

\subsubsection{Fiducial tag localization accuracy}

We calculated tag localization errors for sequences where global registration succeeded. For RANSAC and Teaser, assuming the perfect tag localization accuracy is given on the reconstructed model, we calculated transformation errors of point cloud registration results as fiducial tag localization errors.

Table \ref{tab:result_accuracy} summarizes the fiducial tag localization accuracy of each method. We can see that Teaser exhibited better localization errors (0.180 m and 2.807\textdegree) than that of RANSAC (0.416 m and 7.847\textdegree) thanks to its robust feature matching mechanism. The proposed method showed the best localization accuracy among compared methods (0.110 m and 1.870\textdegree) owing to the robust tag-plane matching. With the refinement step, the localization accuracy was further improved, and we achieved an average translation error of a few centimeters (0.039 m and 1.021\textdegree).

\begin{table*}[tb]
  \centering
  \caption{Fiducial tag localization errors}
  \label{tab:result_accuracy}
  \begin{tabular}{c|cccc}
  \toprule
  \multirow{2}{*}{Method} & 
  \multirow{2}{*}{RANSAC \cite{Buch2013}} &
  \multirow{2}{*}{Teaser \cite{Yang2021}} &
  \multicolumn{2}{c}{Proposed} \\
                                 &                   &                   & w/o refinement    & w/ refinement            \\ \midrule
  Translational error [m]        & 0.416 $\pm$ 0.214 & 0.180 $\pm$ 0.190 & 0.110 $\pm$ 0.078 & {\bf 0.039 $\pm$ 0.060}  \\
  Rotational error [\textdegree] & 7.847 $\pm$ 4.081 & 2.807 $\pm$ 3.042 & 1.870 $\pm$ 1.629 & {\bf 1.021 $\pm$ 1.629}  \\
  \bottomrule
  \end{tabular}
\end{table*}

\subsubsection{Robustness to outliers}

To evaluate the robustness of the proposed method to outlier fiducial tags that are not lying on a plane, we evaluated the global registration success rate while changing the number of inlier tags. For each inlier tag rate setting $R^\text{in}$, we generated $100 R^\text{in}$ fiducial tags on randomly selected planes in the environment and $100  (1 - R^\text{in})$ tags with random poses, and repeated random tag placement and global registration for 100 times. To see how the global registration success rate changes depending on tag relative pose estimation errors, we added two levels of random pose noise ($\sigma_t = 0.05$m / $\sigma_r = 1.0$\textdegree, and $\sigma_t = 0.2$m / $\sigma_r = 4.0$\textdegree) to the tag poses.

\begin{figure}[tb]
    \centering
    \includegraphics[width=\linewidth]{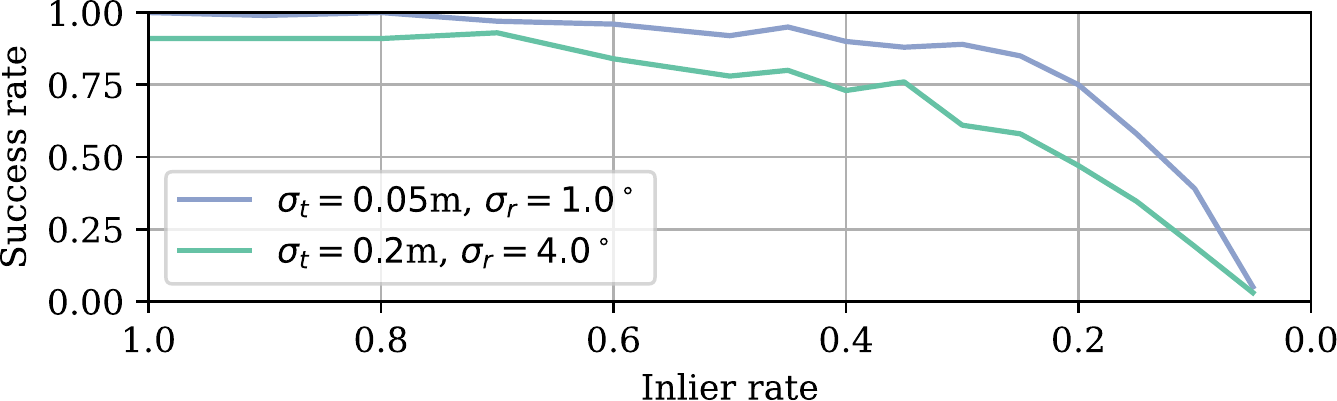}
    \caption{Fiducial tag inlier rate vs global registration success rate. The proposed method is robust to outlier tags that do not lie on a plane.}
    \label{fig:outlier_rate}
\end{figure}

Fig. \ref{fig:outlier_rate} shows a plot of global registration success rate vs inlier rate. We can see that the proposed method is robust to outlier tags and achieved a success rate of over 90\% with 60\% outlier tags under a low-level noise (0.05 m and 1.0\textdegree). Even under larger tag pose noise (0.2 m and 4.0\textdegree), the proposed method achieved a success rate about 75\% with 60\% outlier tags.

\subsection{Evaluation in a real environment}

\begin{figure}[tb]
    \centering
    \includegraphics[width=\linewidth]{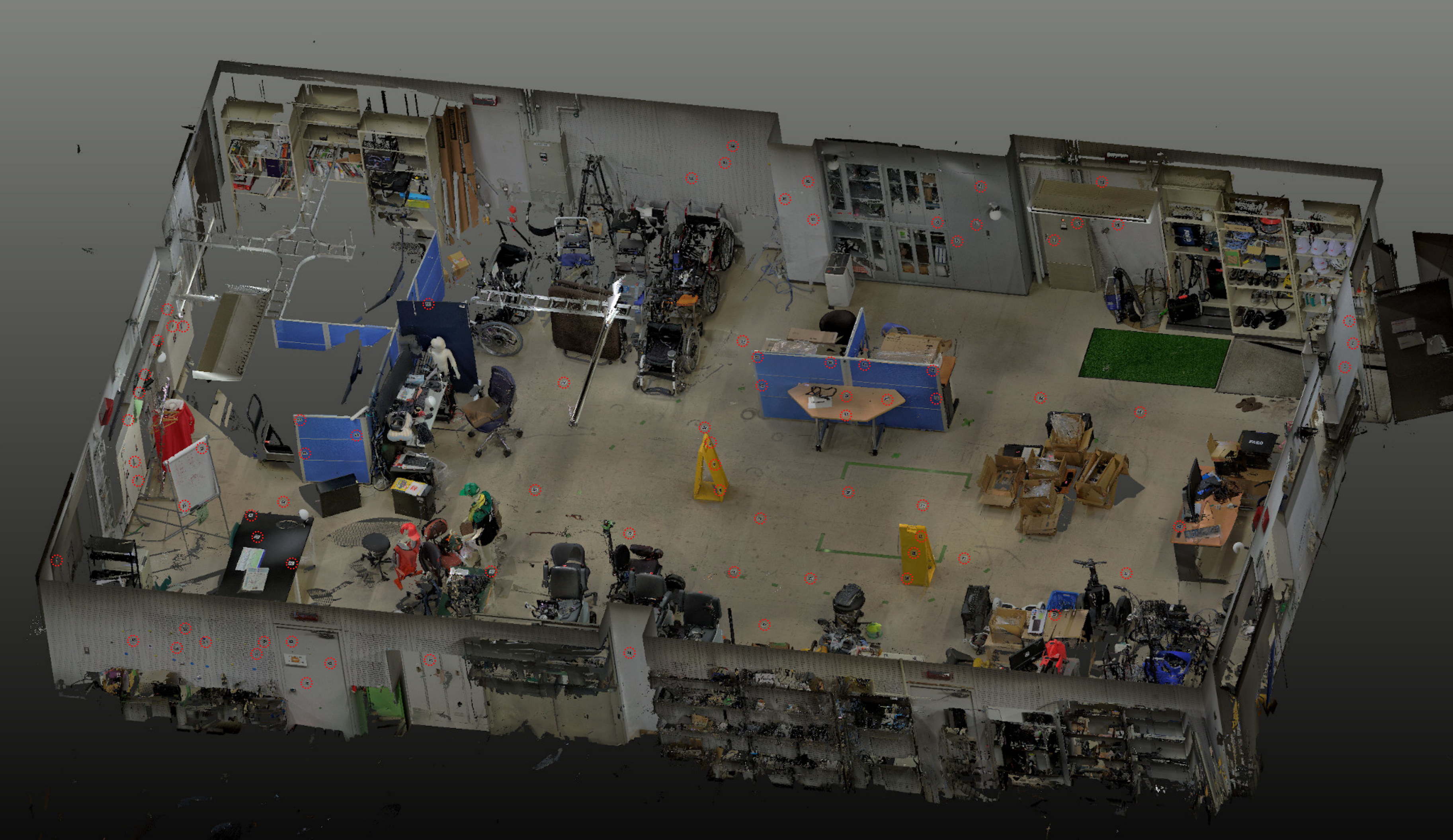}
    \caption{Experimental environment. The red circles indicate the positions of fiducial tags. 117 fiducial tags were placed in the environment.}
    \label{fig:e232}
\end{figure}

To demonstrate that the proposed method enables robust fiducial tag localization on a 3D prior map with a small effort, we placed 117 fiducial tags in the environment shown in Fig. \ref{fig:e232}. The red circles in the figure show the positions of placed fiducial tags. Note that only tags that are visible in the figure are drawn just for visualization. We recorded two environmental map point clouds with and without fiducial tags using a 3D LiDAR (FARO Focus). We manually annotated fiducial tag positions on the environmental map to obtain the ground-truth tag poses. We then estimated the tag poses with the proposed method using the environmental map without tags.

For VIO, we recorded a stream of monocular images and IMU measurements using a MYNTEYE camera. Each fiducial tag was observed by the agile camera at least once during the recording. The duration of the image stream was about 970 s.

Table \ref{tab:proctime} summarizes the processing time of each step in the proposed method. The tag relative pose estimation step was performed on the fly while recording the image stream. The global tag-map registration step took only about 1.5 s thanks to the efficient maximum clique finding algorithm. The refinement step took about 381.3 s in total, and the NID-based camera pose estimation was the most computationally demanding process in this step (379.6 s). Note that the current implementation uses only a CPU, and the processing time of the NID optimization can be improved by 10 to 50 times faster by using a GPU implementation \cite{Oishi2020,Pascoe2015}, resulting in reducing the total processing time to about 10 s. Furthermore, because we can independently perform the image alignment frame-by-frame, it can easily be accelerated by using, e.g., cloud computing.

\begin{figure}[tb]
    \centering
    \includegraphics[width=\linewidth]{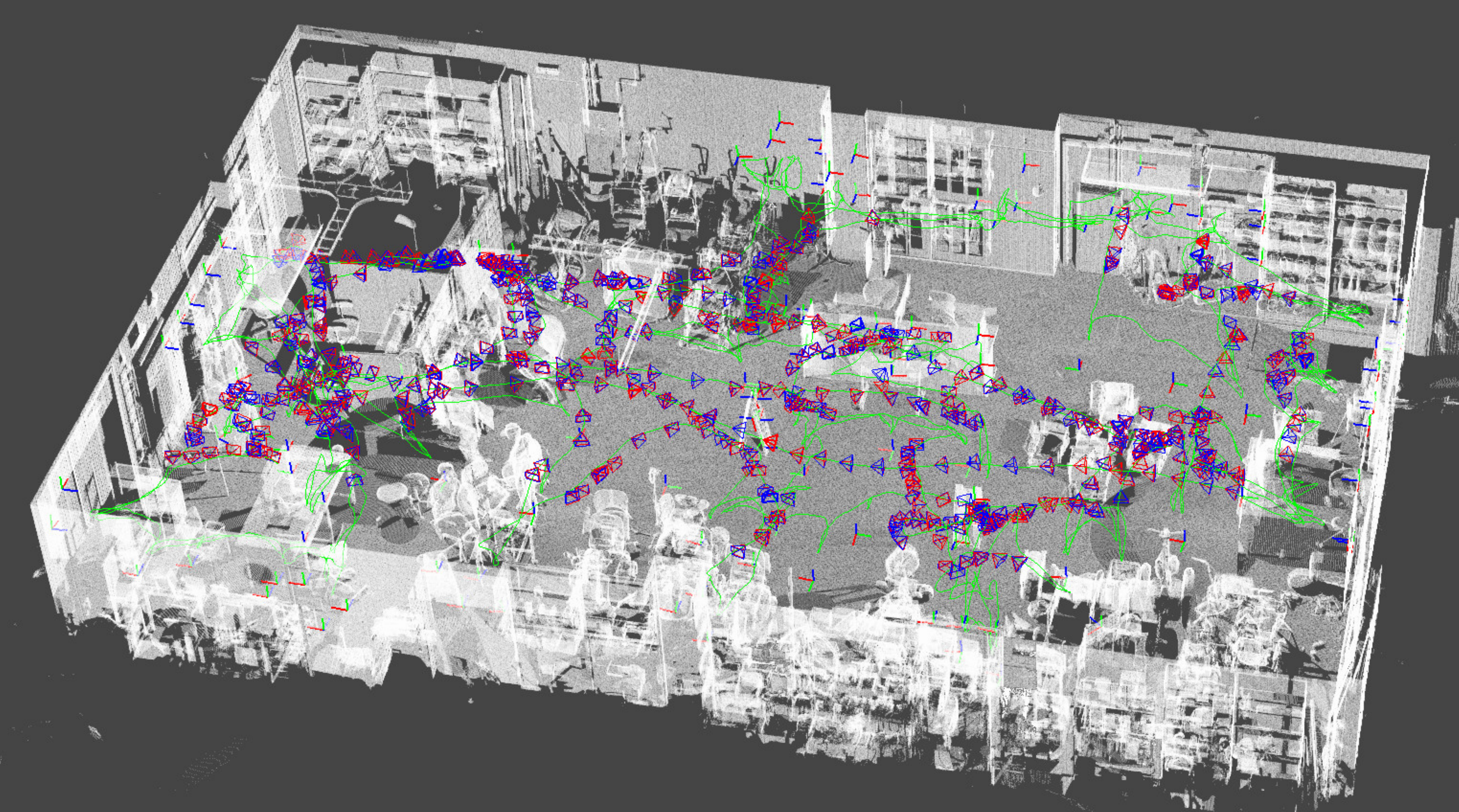}
    \caption{Estimation result. RGB thick lines: estimated tag poses, Green thin line: camera trajectory, Red frustums: camera poses where the NID optimization converged, Blue frustums: NID-based camera-map alignment results.}
    \label{fig:e232_est}
\end{figure}

Fig. \ref{fig:e232_est} shows the estimation result. The thick RGB lines indicate the estimated fiducial tag poses, and the thin green line shows the estimated camera trajectory. The average translational and rotational errors of the estimated fiducial tag poses were respectively 0.019 $\pm$ 0.014 m and 2.382 $\pm$ 4.093 \textdegree. This result would be sufficiently accurate for the requirement for vision-based navigation systems.

%

\begin{table}[tb]
  \centering
  \caption{Processing time}
  \label{tab:proctime}
  \begin{tabular}{lll}
  \toprule
  Step                            & Process                     & Time [s] \\ \midrule
  \multirow{3}{*}{Tag pose estimation}
    & Visual inertial odometry    & \multirow{3}{*}{on-the-fly} \\
    & Fiducial tag detection      &  \\
    & Pose graph optimization     &  \\ \midrule
  \multirow{4}{*}{Global registration}
    & Consistency graph creation  & 1.395 \\
    & Maximum clique finding      & 0.092 \\
    & Transformation optimization & 0.004 \\ \cmidrule{2-3}
    & Total                       & 1.491 \\ \midrule
  \multirow{4}{*}{Estimation refinement}
    & NID camera alignment        & 379.6 \\
    & Outlier filtering           & 0.072 \\
    & Pose graph optimization     & 1.614 \\ \cmidrule{2-3}
    & Total                       & 381.3 \\
  \bottomrule
  \end{tabular}
\end{table}

\section{Conclusions}

We have proposed an accurate and scalable method for fiducial tag localization on a 3D prior environmental map. We first estimate the relative poses between fiducial tags using VIO-based landmark graph SLAM, and then roughly align the fiducial tags with a 3D prior map using a graph-theoretic tag-plane correspondence estimation. We refine the estimated tag and camera poses by directly aligning camera images with the environmental map using an information-theoretic metric. Through simulation-based experiments, the proposed method achieved a global registration success rate of 98\% and tag estimation accuracy of a few centimeters. The real experiment demonstrated that the proposed method can accurately localize over 100 fiducial tags on a prior map in 16 minutes for data recording and 6 minutes for post-processing.

\section*{Appendix}

\subsection{Tag-Plane Correspondence Consistency Check}
\label{sec:consistency}

Alg. \ref{alg:consistency} describes the algorithm to determine the consistency of two tag-plane correspondence hypotheses. ${\bf n}_*, R_*, {\bf p}_*$ are the normal, rotation, and translation of a tag or plane, respectively. ${\bf l}_*$ is the length of a plane along XYZ axes (Z = normal). Lines 2 and 3 swap $h_{ij}$ and $h_{kl}$ if the normal of $p_j$ is almost vertical to avoid indeterminacy of the tag-plane rotation (Because we can assume that both $t_i$ and $p_j$ are gravity-aligned, the tag-plane rotation can be determined from only their normals as long as the plane normal is not vertical). Lines 4-7 calculate the transformation that transforms $t_i$ such that its normal is aligned with the normal of $p_j$. Note that we consider only rotation along the gravity direction assuming tags and planes are gravity-aligned. Line 8 checks if the angle error between the normals of $p_l$ and rotated $t_k$. If it is larger than a threshold, we consider $h_{ij}$ is inconsistent with $h_{kl}$. Lines 10-13 calculate the translation to minimize the distance between $t_k$ and $p_l$ such that $t_i$ remains on $p_j$. Then, if the distance between transformed $t_k$ and $p_l$ is smaller than a threshold, we consider $h_{ij}$ and $h_{kl}$ are mutually consistent.


\begin{algorithm}[tb]
\caption{Tag-plane correspondence consistency check}
\label{alg:consistency}
\begin{algorithmic}[1]
  \Function{consistency\_check}{$h_{ij}, h_{kl}$}
    \If{${\bf n}_j \cdot [0, 0, 1]^T > 1 - \epsilon$}
      \State{\Call{swap}{$h_{ij}, h_{kl}$}}
    \EndIf

    \State ${\bf n}_i^{\text{XY}} = {\bf n}_{t_i} \circ [1, 1, 0]^T$
    \State ${\bf n}_j^{\text{XY}} = {\bf n}_{p_j} \circ [1, 1, 0]^T$
    \State $R_{ji} = \Call{align\_vectors}{{\bf n}_i^{\text{XY}}, {\bf n}_j^{\text{XY}}}$
    \State ${\bf p}_{ji} = {\bf p}_{p_j} - R_{ji} {\bf p}_{t_i}$

    \If{$\Call{Angle}{R_{ji} {\bf n}_{t_k}, {\bf n}_{p_l}} > \text{th}^\text{rot}$}
        \State \Return{False}
    \EndIf

    \State ${\bf p}_{lk} = {\bf p}_{p_l} - (R_{ji} {\bf p}_{t_k} + {\bf p}_{ji})$
    \State ${\bf p}_{jk} = R_{p_j}^{-1} {\bf p}_{lk}$
    \State ${\bf p}_{jk}' = \Call{Clamp}{{\bf p}_{jk}, -{\bf l}_{p_j} / 2, {\bf l}_{p_j} / 2}$
    \State ${\bf p}_{t_k}' = R_{ji} {\bf p}_{t_k} + {\bf p}_{ji} + R_{p_j} {\bf p}_{jk}'$

    \If{$\Call{Distance}{{\bf p}_{t_k}', p_l} > \text{th}^\text{trans}$}
        \State \Return False
    \Else
        \State \Return True
    \EndIf

  \EndFunction

  \Function{align\_vectors}{${\bf a}, {\bf b}$}
    \State ${\bf v} = {\bf a} \times {\bf b}$
    \State $s = \| {\bf v} \|$
    \State $c = {\bf a} \cdot {\bf b}$
    \If{$s < \epsilon$}
        \State{\Return $I$}
    \EndIf
    \State $S = \text{skew}({\bf v})$
    \State \Return $I + S + \frac{1-c}{s^2} S^2$
  \EndFunction
\end{algorithmic}
\end{algorithm}

\balance

\bibliographystyle{IEEEtran}
\bibliography{iros2022}

\end{document}